\documentclass[journal]{IEEEtran}

\usepackage{url}

\usepackage{svg}
\usepackage{booktabs}
\usepackage{multirow}
\usepackage{xspace}
\usepackage{amsmath}

\newcommand{\mnamens}{PhISM} 
\newcommand{\mname}{\mnamens\xspace}

\DeclareMathOperator\erf{erf}

\begin{document}

\title{Physics-Informed Spectral Modeling\\
for Hyperspectral Imaging}
%
%
%

\author{Zuzanna~Gawrysiak
        and~Krzysztof~Krawiec
\thanks{Z. Gawrysiak and K. Krawiec are with the Institute
of Computing Science, Poznan University of Technology, Poznan, Poland
e-mail: \{zgawrysiak, kkrawiec\}@cs.put.poznan.pl.}
\thanks{Manuscript received xxxx; revised September xxxx.}}

\markboth{IEEE Geoscience and Remote Sensing Letters,~Vol.~xx, No.~xx, xxxx}%
{Shell \MakeLowercase{\textit{et al.}}: Bare Demo of IEEEtran.cls for Journals}
%



\maketitle

\begin{abstract}
We present \mname, a physics-informed deep learning architecture that learns without supervision to explicitly disentangle hyperspectral observations and model them with continuous basis functions. \mname outperforms previous methods on several classification and regression benchmarks, requires limited labeled data, and provides additional insights thanks to its interpretable latent representation.   
\end{abstract}


%
\IEEEpeerreviewmaketitle


\section{Introduction}

\IEEEPARstart{H}{yperspectral} remote sensing (RS) captures a high-resolution signature, a physically grounded pattern that describes how materials reflect or absorb light across wavelengths, and enables fine-grained discrimination that far exceeds the capabilities of conventional imaging.
However, the high-dimensional feature space challenges machine learning (ML) methods: models require more parameters, become data-hungry, and prone to overfitting, especially when labels are scarce. Conventional deep learning (DL) models, such as Convolutional Neural Networks (CNNs), treat input channels as independent features, ignoring the physical correlations between neighboring spectral bands, which forces the model to re-discover the well-established knowledge from data alone and increases the risks of forming hypotheses that are physically implausible \cite{geo-aware,dl-rs}. 

To address these challenges, we propose \emph{Physically Informed Spectral Modeler} (\mname), an architecture that incorporates domain knowledge by representing spectral components with transparent latent basis functions, 
each controlled by a small number of interpretable parameters. \mname achieves performance competitive with state-of-the-art techniques, works robustly with limited training data, and, by operating in the realm of spectral features that are familiar to geoscientists, is more interpretable than DL methods.



\section{The approach}\label{sec:method}

\mname is based on the autoencoder blueprint and involves two stages (Fig.\ \ref{fig:autoencoder}): (i) autoassociative, self-supervised and task-agnostic training of the autoencoder, to form informative latent representations that enable possibly accurate reconstruction of the input image (Section~\ref{sec:reconstruction}), and (ii) task-specific training of a \emph{prediction module} that maps that latent representation to the respective dependent variables (Section~\ref{sec:prediction}). 

\subsection{Self-supervised modeling and reconstruction of spectra} \label{sec:reconstruction}

\begin{figure}[t!]
    \centering
    \includegraphics[width=\linewidth]{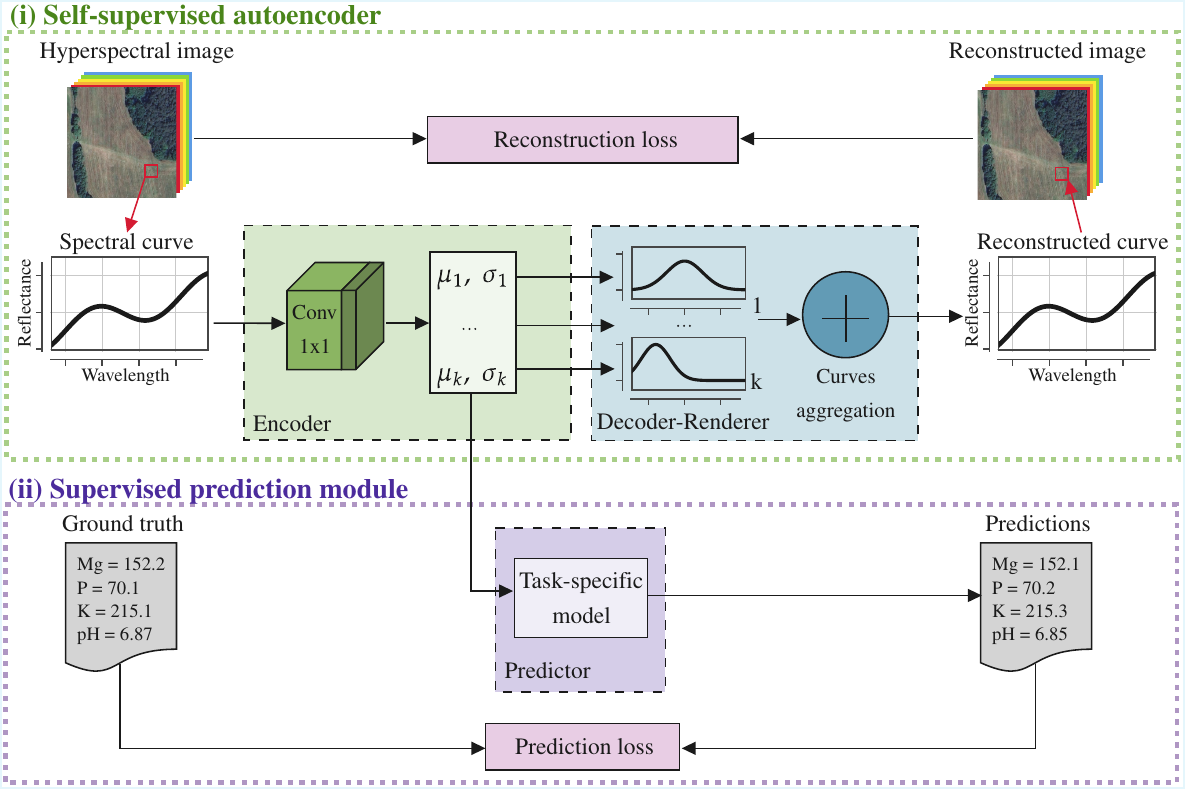}
    \caption{Self-supervised training of \mname (top) and its application to supervised classification and regression tasks (bottom).}
    \label{fig:autoencoder}
\end{figure}

\mname's autoencoder comprises an \emph{encoder} and a \emph{decoder} that communicate via compact latent representation (Fig.\ \ref{fig:autoencoder}). The model processes each image pixel independently, in parallel, which ensures that the learned estimates rely exclusively on the physical spectral signature and avoid confounding spectral signals with spatial textures and information leakage from the neighborhood. 
The encoder is a lightweight, pixel-wise CNN feature extractor comprising five 1 × 1 convolutional layers with 512, 1024, 512, 256, and 4$k$ channels, respectively ($\sim$1.2M total parameters), each followed by batch normalization and Leaky ReLU activation. The input dimension matches the number of spectral bands. 
The decoder, rather than relying on typical DL components, explicitly parametrizes $k$ continuous spectral components represented with \emph{basis functions}, which together form the reconstructed spectrum. The decoder is thus `expressing' the $k$ spectral components parameterized by the encoder; since this comes down to \emph{sampling} of the basis functions at specific wavelengths, we refer to it as \textit{renderer}.


For the model to be trainable end-to-end with gradient, the basis functions need to be differentiable, which holds for, e.g., splines, polynomials, normal distributions, beta distributions, and skew normal distributions. Here, we use the skew normal distribution, as it turned out to fare best in preliminary experiments on the HYPERVIEW benchmark (Sec.~\ref{sec:results}) in terms of reconstruction measured by PSNR (splines: 40.5, polynomials: 38.3, normal: 38.8, beta: 26.5, skew normal: 48.5). Its probability density function (PDF) $f$ is parametrized by the mean ($\mu$), standard deviation ($\sigma$), and skew ($\alpha$):  
\begin{equation}
f(\lambda \mid \mu, \sigma, \alpha) = 2 \mathcal{N}(\lambda \mid \mu, \sigma) F_{\mathcal{N}}(\alpha (\lambda - \mu)),\label{eq:skew}
\end{equation}
where $\lambda$ is the wavelength, $\mathcal{N}(\lambda|\mu, \sigma)$ is the PDF of the normal distribution and $F_{\mathcal{N}}(\lambda)$ is the cumulative distribution function (CDF) of the standard normal distribution,  
$F_{\mathcal{N}}(\lambda) = \frac{1}{2} \left[1 + \erf (\lambda/\sqrt{2})\right]$. 
The total estimate at wavelength $\lambda$ is:  
\begin{equation}
S(\lambda) = \sum_{i=1}^k s_i f(\lambda \mid \mu_i, \sigma_i, \alpha_i), \label{eq:s}
\end{equation}
where $\mu_i$, $\sigma_i$, and $\alpha_i$ parameterize the $i$th spectral component, and the scale $s_i$ modulates its contribution to the spectrum. To calculate the per-pixel output, the renderer simply queries $S(\lambda)$ at the wavelengths $\lambda$ corresponding to the input bands. 

There are thus $4$ parameters per spectral component ($\mu_i$, $\sigma_i$, $\alpha_i$ and $s_i$), which requires $4k$ dimensions in the latent, 
 where $k$ is usually moderate ($\leq10$). We use the sigmoid activation function for $\mu_i$ and $\sigma_i$ to ensure non-negativity, and the $\tanh$ activation function for $\alpha_i$ and $s_i$. Then, the outputs of the activation functions are multiplied by the number of spectral channels (e.g. 224 for the AVIRIS sensor) and fed into Eq.\ (\ref{eq:s}). Notice that the signed $s_i$ allows the components to contribute positively or negatively. The model works with spectra that are zero-centered w.r.t. the means calculated from the training set, i.e. Eq.\ (\ref{eq:s}) estimates the signed divergence from them. 

Training follows the standard autoencoder blueprint: the encoder produces the latent vector, the decoder uses it to render and combine the spectral components, and the resulting spectrum is compared to the input spectrum with the Huber loss function~\cite{huber1964}, which combines the advantages of MSE and MAE. The AdamW optimizer~\cite{adamw} updates the encoder's parameters (the renderer has no trainable parameters) at 0.0001 learning rate for 50 epochs or until the validation loss does not improve for 5 epochs (early stopping). The training process is \emph{entirely self-supervised}, and thus does not require ground-truth data, which is often scarce and hard to come by. 

To select the optimal number of components, we performed a sensitivity analysis by measuring the PSNR for $k \in [1, 15]$ and observed that it increases from 42.1 at $k=1$ to 48.5 at $k=5$, and then plateaus. Consequently, we use $k \in [5,10]$ to maximize interpretability without compromising fidelity. 

Though the composition of separately modeled spectral components bears resemblance to spectral unmixing \cite{spectral-unmixing-th}, \mname significantly diverges from it by (i) not relying on predefined spectral components, but learning them from data, and (ii) modeling them with smooth basis functions, to match the characteristics and variability of spectral patterns, while keeping their complexity at bay. Rather than aiming at maximally faithful modeling of physical processes, we aim at a degree of physical plausibility that both \emph{constrains} and \emph{informs} our models, so that they generalize well.

\subsection{Supervised learning for prediction of dependent variables} \label{sec:prediction}

Once the autoencoder has been trained, we discard the renderer (decoder) and use the compact interpretable latent features for predictive downstream regression and classification tasks. We achieve this by appending an arbitrary ML model to the encoder and training it in a supervised fashion on the available labeled ground-truth data (bottom part of Fig.~\ref{fig:autoencoder}). Because the number of latent features is low, well-performing predictive models can be trained even from very small samples of labeled pixels (Sec.\ \ref{sec:results}). Also, one can opt for a transparent ML model (e.g., a decision tree) to improve the overall interpretability.

\section{Related work}

Incorporating domain-specific knowledge~\cite{domain-dl} and physical principles~\cite{physics-informed} into ML models bridges data-driven and physically-grounded approaches, enabling better generalization and interpretability. The efficacy of enforcing physical constraints has been well-established in many engineering domains, e.g. thermal~\cite{battery-thermal} and electrochemical modeling~\cite{battery-electro}.
In RS, a range of works attempted to inject the relevant priors into DL models explicitly, e.g. by engaging predefined ontologies \cite{ontology-rs}. 
Zheng et al.\ \cite{spectral-unmixing} combined spectral unmixing with deep learning to enhance image fusion and generate high-resolution hyperspectral images from high-resolution multispectral and low-resolution hyperspectral inputs. Unsupervised dehazing networks augmented with hybrid priors have shown promising results in improving the quality of hyperspectral images~\cite{dehazing-prior}. 

Physics-inspired approaches also provide robust solutions for unsupervised super-resolution of hyperspectral data, as demonstrated by the physics-driven autoencoder presented by \cite{model-autoencoder}. Camps-Valls et al.\ \cite{physics-geoscience} integrated physics-driven insights to address geoscience-specific challenges in RS. CRANN \cite{crann} used physics-based principles combined with neural networks to retrieve cloud properties from hyperspectral measurements. Li et al.\ \cite{toblers-law} leveraged spatial autocorrelation to explicitly account for spatial relationships, enabling improved detection of terrain features under weak supervision. VarioCNN~\cite{glacier} combined physically constrained neural networks with deep CNNs to analyze complex glaciological processes (crevasse classification). GASlumNet~\cite{slums} integrated DL with geoscientific prior knowledge to improve slum mapping accuracy. Ge et al.\ \cite{geo-aware} outlined the Geoscience-Aware DL paradigm that integrates geoscience knowledge into DL frameworks at various stages of modeling. These methods leveraged domain physics \emph{extrinsically}, via simulated training data~\cite{crann} or spatial statistics~\cite{glacier}. In contrast, \mname embeds physics \emph{intrinsically} by embodying continuous basis functions (Eq.~(\ref{eq:s})).

Given that the RS-specific domain knowledge can be often represented in symbolic form, a number of works can be seen as subscribing to the paradigm of \emph{neurosymbolic} AI \cite{neural-symbolic, neurosymbolic-ai}. Harmon et al.\  \cite{tree-crown} used probabilistic soft logic rules to encode expert insights into a neuro-symbolic model, improving tree crown delineation and enabling generalization beyond annotated data.
Incorporating domain knowledge in the form of equations embedded in the loss function proved particularly effective in the classification of tree species, while also enhancing explainability \cite{trees-neuro}. Chen et al.\ \cite{neurosymbolic-minerals} discussed implications for mineral prediction, underscoring the synergy between symbolic reasoning and neural methods. Potnis et al.~\cite{neuro-scene} integrated geospatial knowledge graphs into DL models to enhance neurosymbolic AI for RS scene understanding.

\mname's novelty in relation to past work consists in explicit modeling of spectral components using continuous, differentiable formulas, which facilitates self-supervised training from small data and is more interpretable than DL approaches. 

\begin{table}[t]
    \centering
    \caption{Average predictive accuracy ($\uparrow$) on classification tasks with $.95$ confidence intervals (results for BAAS cited from \cite{nalepa2020igarss})}
    \begin{tabular}{lcccccc} 
    \toprule
        \multirow{2}{*}[-2pt]{Method} 
        & \multicolumn{2}{c}{Salinas Valley} 
        & \multicolumn{2}{c}{Pavia University} 
        & \multicolumn{2}{c}{Indian Pines} \\
        \cmidrule(lr){2-3} \cmidrule(lr){4-5} \cmidrule(lr){6-7}
        & OA & AA & OA & AA & OA & AA \\
    \midrule
        3D~\cite{3d-cnn} & 69.7 & 69.1 & 70.1 & 60.2 & 48.9 & 38.3 \\
        1D~\cite{splits-hsi} & 64.2 & 64.7 & 73.3 & 62.1 & 67.1 & 55.1 \\
        BAAS~\cite{7938656} & 73.4 & 74.3 & 69.5 & 60.4 & 46.8 & 35.4 \\
        SF~\cite{hong2021spectralformer} & 68.1{\scriptsize$\pm$3.2} & 67.7{\scriptsize$\pm$2.4} & 69.9{\scriptsize$\pm$1.2} & 59.1{\scriptsize$\pm$1.6} & 48.6{\scriptsize$\pm$0.6} & 39.5{\scriptsize$\pm$1.1} \\
        3DAES~\cite{3daes} & 73.1{\scriptsize$\pm$2.8} & 77.8{\scriptsize$\pm$2.2} & 68.5{\scriptsize$\pm$1.5} & 69.2{\scriptsize$\pm$1.7} & 63.7{\scriptsize$\pm$0.5} & 53.1{\scriptsize$\pm$1.0} \\
        Autoencoder & 71.4{\scriptsize$\pm$3.8} & 76.2{\scriptsize$\pm$2.3}  & 66.1{\scriptsize$\pm$1.8} & 66.5{\scriptsize$\pm$1.1} & 59.8{\scriptsize$\pm$0.7} & 50.4{\scriptsize$\pm$0.8} \\
        \mname (ours) & 73.4{\scriptsize$\pm$3.8} & 78.3{\scriptsize$\pm$2.5} & 67.4{\scriptsize$\pm$1.9} & 68.0{\scriptsize$\pm$1.2} & 64.4{\scriptsize$\pm$0.4} & 54.6{\scriptsize$\pm$0.8}\\
        \mname (fixed) & 70.3{\scriptsize$\pm$4.0} & 75.1{\scriptsize$\pm$2.7} & 66.1{\scriptsize$\pm$1.9} & 66.6{\scriptsize$\pm$1.3} & 57.7{\scriptsize$\pm$0.5} & 48.8{\scriptsize$\pm$0.9}\\
    \bottomrule
    \end{tabular}
    \label{tab:classification}
    \vspace{-3mm}
\end{table}

\section{Results}\label{sec:results}

We demonstrate \mname on a number of classification and regression benchmarks, following the procedure outlined in Sec.\ \ref{sec:method}: we fit the autoencoder to the training set (Sec.~\ref{sec:reconstruction}) and combine it with a predictive ML model, which we train to map the encoder's latent to the dependent variable (Sec.\ \ref{sec:prediction}).
The data is first zero-centered by decreasing the values in each spectral band by the average calculated from the training set. 
 
The method has been implemented in PyTorch. A typical cross validation experiment took, respectively, $\sim$8 and $\sim$30 minutes for a single classification and regression benchmark, on an NVIDIA A100 GPU with 80 GB of VRAM.
Technical details can be found in the source code repository.\footnote{\url{https://github.com/zuzg/domain-aware-hyperspectral-ml}} 

\subsection{Results for classification tasks}

We use the modernized versions of three popular \emph{pixel classification} benchmarks:
    Salinas Valley (\textbf{SV}), agricultural area captured with the AVIRIS sensor\footnote{\url{https://aviris.jpl.nasa.gov}}, 224 bands, 16 classes;
    Pavia University (\textbf{PU}), urban area captured with ROSIS sensor, 103 bands, 9 classes;
    Indian Pines (\textbf{IP}), mixed agricultural/forest area, AVIRIS sensor, 200 bands, 16 classes.
To avoid information leakage and provide fair and reproducible comparison, we use the fixed partitioning of data into a training part (spatially disjoint patches) and testing parts (all remaining pixels) proposed in~\cite{splits-hsi}.\footnote{
Random partitioning of pixels into training and test sets leads to information leaks and overly optimistic accuracy estimates, up to 100\%~\cite{feng2023}.} In each of 4 (IP) or 5 (SV, PU) cross-validation folds, we first train our autoencoder with $k=10$ spectral components (Sec.\ \ref{sec:reconstruction}), resulting in a $4k=40$-dimensional latent representation. These features are then used to train a pixel-wise XGBoost classifier \cite{xgboost} (Sec.\ \ref{sec:prediction}). We report the overall accuracy \textbf{(OA)}, i.e. the ratio of the correctly predicted pixels over all test pixels, and the average accuracy \textbf{(AA)}, i.e. the mean of per-class accuracies, to address the class imbalance. We repeated the training and testing in each fold 5 times with different seeds, so the presented results summarize 20 (IP) or 25 (SV, PU) runs of the method.

\begin{figure}[t!]
    \centering
    \includegraphics[width=\linewidth]{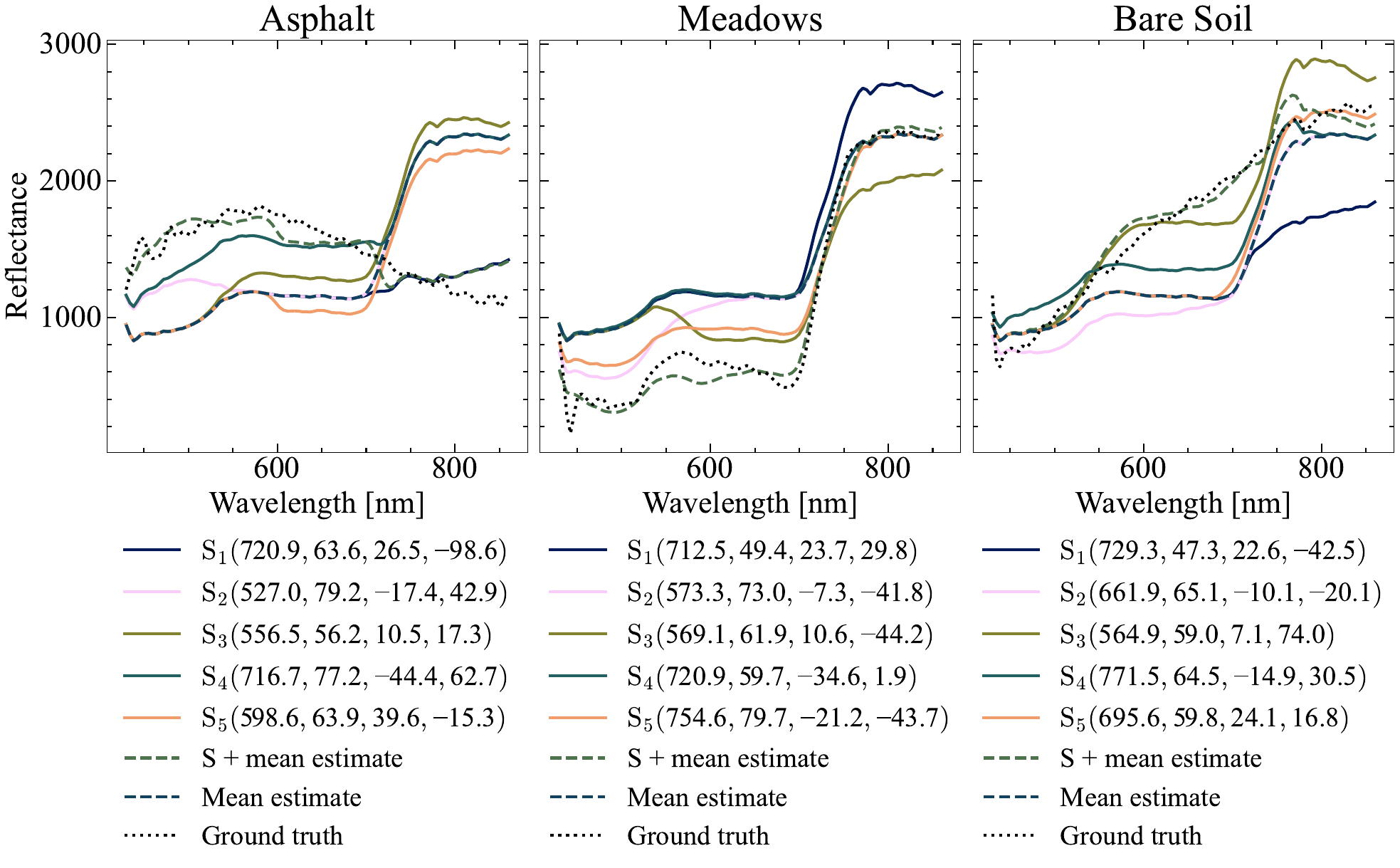}
    \caption{Parametric variation of spectral components, illustrated by querying the model on random pixels from three decision classes from PU dataset. $S_i(\mu_i,\sigma_i,\alpha_i, s_i)$ denotes the parameters of $i$th component in Eq.\ \ref{eq:s}. The plots present the values of $S_i$ added to the mean estimated for zero-centering. }
    \label{fig:spectral-curves}
    \vspace{-3mm}
\end{figure}

\subsubsection{Results}
In Table \ref{tab:classification}, we compare \mname against six methods: \textbf{1D CNN}~\cite{splits-hsi}, operating on per-pixel spectra, \textbf{3D CNN}~\cite{3d-cnn}, processing small spatial-spectral cubes, Band-Adaptive Spectral-Spatial Feature Learning, \textbf{BAAS}~\cite{7938656}, \textbf{SpectralFormer (SF)}~\cite{hong2021spectralformer}, a Transformer-based architecture, \textbf{3DAES}~\cite{3daes}, an autoencoder-based architecture, and the conventional DL autoencoder. The latter comprises the same encoder architecture as \mname's and a 1×1-convolutional decoder that `mirrors' the encoder (doubling thus the number of \mname's parameters); the XGBoost learns from the $4k$-dimensional latent of this model. To ensure a fair comparison, the autoencoder was tuned using Optuna~\cite{optuna}.

Despite not being optimized specifically for segmentation, \textbf{\mname} achieves the best AA on SV and PU, and is competitive on IP, confirming the generality and strong discriminative capacity of the learned representations. On OA, \mname yields to other methods; however, this metric largely neglects the smaller decision classes, which is particularly undesirable for the considered benchmarks, where the number of pixels per decision class can vary by more than an order of magnitude.

In the ablated \textbf{\mname (fixed)} variant, $\mu_i$, $\sigma_i$ and $\alpha_i$ are optimized in training, but do not depend on the observed input spectrum (like biases in DL units). These models form \emph{fixed} spectral components that are mixed linearly with the input-dependent scales $s_i$, akin to spectral unmixing (cf. Sec.\ \ref{sec:method}). The significantly worse performance of this variant corroborates the need for pixel-wise shaping of spectral components. 

\subsubsection{Visualization of components}
Figure~\ref{fig:spectral-curves} presents the $k=5$ spectral components produced by one of the models trained on PU for three testing pixels selected randomly from the largest decision classes: \emph{asphalt}, \emph{meadows}, and \emph{bare soil}. Curve color corresponds to component index ($i$ in Eq.\ (\ref{eq:s})). In contrast to spectral unmixing that controls only the weights of spectral components, \mname also modulates their shapes and can model both the positive and the negative contributions, which in principle allows capturing, respectively, emission and absorption at particular wavelengths. 

\subsubsection{Interpretability}The explicit representation of components eases interpretation of inference conducted by \mname. For instance, the parameters of $S_i$ shown for the example pixels in Fig.~\ref{fig:spectral-curves} reveal that 
consecutive components tend to focus on increasing wavelengths, with $S_1$ operating around the green hue, while $S_5$ covering infrared wavelengths. Further insights can be obtained by, e.g., inspecting attribute importance using the Shapley interaction values~\cite{shapiq}.

Each encoder instance, by starting training from a random initial configuration of parameters, may in principle converge to different spectral components. 
To assess the replicability of this process, we trained and evaluated 10 models and examined the distributions of the four predicted parameters of skew normal functions.  The median of per-image standard deviations for $\mu$ and $\sigma$ ($[0,1]$ range) were below 0.04, and below 0.09 for $\alpha$ and $s$ ($[-1,1]$ range). This stability shows that representation biases imposed by the skew normal functions and the low-dimensional latent space regularize the model effectively.

\subsubsection{Emergence of structure in the latent}
\begin{figure}[t!]
    \centering
    \includegraphics[width=.85\linewidth]{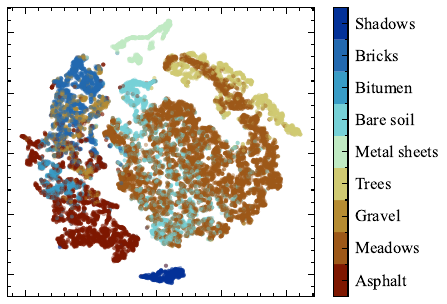}
    \caption{t-SNE~\cite{tsne} projection of training data for PU.}
    \label{fig:projection}
    \vspace{-3mm}
\end{figure}

Figure~\ref{fig:projection} presents the 2D projection of \mname's latent space, obtained by applying the t-SNE method~\cite{tsne} to the 20 parameters that control the $k=5$ spectral components in the model trained on the PU dataset. 
Clusters of observations that represent materials of similar constitution (e.g., \emph{Bitumen} and \emph{Asphalt}, \emph{Meadows} and \emph{Bare\ soil}) tend to overlap, which suggests that self-supervision was sufficient to adequately capture their spectral similarity. Conversely, classes that have little in common (e.g., \emph{Asphalt} and \emph{Meadows}) are clearly separated. Some classes (\emph{Metal sheets}, \emph{Shadows}) form compact, isolated clusters, which in principle allows delineating them without explicit labeling of pixels (i.e., labeling them post-hoc).   


\subsubsection{Learning from small data}
\begin{table}[t!]
    \centering
    \caption{Test-set AA of models trained on smaller PU training sets.}
    \begin{tabular}{lccccc}
    \toprule
        \multirow{2}{*}[-2pt]{Method} 
        & \multicolumn{5}{c}{Train-set percentage} \\
        \cmidrule(lr){2-6}
        & 50\% & 10\% & 5\%& 1\%& 0.5\%\\
    \midrule
    3DAES~\cite{3daes} & 84.7{\scriptsize$\pm$0.3} & 83.9{\scriptsize$\pm$0.6} & 83.2{\scriptsize$\pm$0.7} & 75.4{\scriptsize$\pm$1.1} & 69.1{\scriptsize$\pm$1.3}\\
         Autoencoder & 79.9{\scriptsize$\pm$0.2} & 77.8{\scriptsize$\pm$0.6} & 76.6{\scriptsize$\pm$0.6} & 67.5{\scriptsize$\pm$2.9} & 65.4{\scriptsize$\pm$1.5}\\
        Raw & 81.3{\scriptsize$\pm$0.4} & 80.8{\scriptsize$\pm$0.5} & 78.6{\scriptsize$\pm$0.8} & 72.9{\scriptsize$\pm$1.9} & 56.1{\scriptsize$\pm$2.5}\\
        \mname (ours) & 82.5{\scriptsize$\pm$0.3} & 80.1{\scriptsize$\pm$0.7} & 79.2{\scriptsize$\pm$0.8} & 73.6{\scriptsize$\pm$1.6} & 70.0{\scriptsize$\pm$3.0}\\
    \bottomrule
    \end{tabular}
    \label{tab:small-data}
    \vspace{-3mm}
\end{table}

To simulate label-scarce conditions, we trained independent XGBoost models on small subsets (0.5-50\%) of the PU training set processed with the same encoder architectures as in Table~\ref{tab:classification}, but trained with $k=5$, and queried them on the fixed set of the remaining 50\% of pixels. The values of AA obtained by repeating this process 10 times for different random seeds, reported in  Table~\ref{tab:small-data}, are higher than in  Table~\ref{tab:classification}, because the partitioning of pixels into train and test sets is here random. Crucially however, \mname fares systematically better than for Raw and Autoencoder and degrades more gracefully when labeled training data become gradually more scarce. 
While the self-supervised 3DAES~\cite{3daes} is less impacted by moderate deprivation of labeled data (10-5\%), \mname maintains comparable performance in the extreme low-data regime (0.5\%). This stability suggests that \mname's physics-informed constraints act as a regularizer.

\subsection{Results for regression tasks}

We apply \mname to the regression tasks posed in the HYPERVIEW challenge~\cite{hyperview} (\textbf{H1}, data acquired with the HySpex VS-725 sensor) and HYPERVIEW 2 challenge (\textbf{H2}, data from PRISMA\footnote{\url{https://directory.eoportal.org/satellite-missions/prisma-hyperspectral}})\footnote{\url{https://platform.ai4eo.eu/hyperview2}}. For H1, the soil parameters to be predicted are K, P, Mg, and pH level; for H2, these are B, Cu, Zn, Fe, S, and Mn.  
In contrast to the above classification tasks, the dependent variables in H1 and H2 are given \emph{per image patch}, rather than per pixel. We use only the publicly available parts from both challenges, for which the values of the dependent variables are available. For H1, these are 1,732 patches, which we divide into 1,000 training samples, 124 for validation, and 608 for testing; the average patch size is 60 × 60 pixels with 150 hyperspectral bands. 
For H2, there are 1,876 patches, which we divide into 1,000 training samples, 124 for validation, and 752 for testing; the average patch size is 2 × 2 pixels (60x60 meters) with 230 hyperspectral bands.

The self-supervised phase of training remains the same as in classification, i.e. the model learns to reproduce the spectrum in each pixel, with $k$ set to 5. We then average the latent representations per patch and train on them a separate Random Forest~\cite{random-forest} regressor for each of the dependent variables. 

Table \ref{tab:regression} compares the performance of \mname to the baselines in terms of the error score used in the challenges (\emph{Hyperview score} \cite{hyperview}), which aggregates the $\text{MSE}_i$ errors committed on all dependent $n$ variables relative to fixed baselines $\text{MSE}_i^\text{base}$ as $\frac{1}{n}\sum_{i=1}^n  (\text{MSE}_i / \text{MSE}_i^\text{base})$. The baselines are simple Autoencoder (as in classification tasks) and Raw configurations, in which the Random Forest learns directly from the spectral channels averaged over a patch. \mname slightly outperforms both baselines on H1; for H2, its superiority is much more evident. The fixed variant fares worse again, confirming the usefulness of the pixel-dependent prediction of all parameters of \mname's spectral components.

\begin{table}[t]
    \centering
    \caption{Average predictive error score ($\downarrow$) on regression tasks}
\begin{tabular}{lcccc}
\toprule
Dataset & Raw & Autoencoder & \mname (ours) & \mname (fixed) \\
\midrule
H1 & 0.723{\scriptsize$\pm$0.066} & 0.732{\scriptsize$\pm$0.069} & 0.721{\scriptsize$\pm$0.064} & 0.798{\scriptsize$\pm$0.091} \\
H2 & 0.501{\scriptsize$\pm$0.098} & 0.493{\scriptsize$\pm$0.091} & 0.389{\scriptsize$\pm$0.095} & 0.484{\scriptsize$\pm$0.081} \\
\bottomrule
\end{tabular}

    \label{tab:regression}
    \vspace{-3mm}
\end{table}


\section{Conclusion}
We have shown that equipping DL models with physics-inspired priors informs them effectively and offers better predictive accuracy, lower demand for labeled data, and more transparency of the inference process. Overall, the neurosymbolic architectures \cite{neural-symbolic, neurosymbolic-ai} offer a particularly promising and natural framework for incorporating the wealth of RS-related domain knowledge, and will continue to be the subject of our further research.
Among others, we plan to exploit \mname's use of continuous physical parameters (e.g., wavelength $\mu$ in nm), rather than discrete band indices, as it facilitates cross-sensor transferability: unlike in standard CNNs, the learned latent representation is sensor-agnostic. Future work will transfer models between sensors by mapping diverse spectral samplings to this unified physical space.

\noindent\textbf{Acknowledgment:} Research supported by the statutory funds of Poznan University of Technology and the Polish Ministry of Science and Higher Education, grant no. 2025/57/B/ST6/03737.

\ifCLASSOPTIONcaptionsoff
  \newpage
\fi



\bibliographystyle{IEEEtran}
\bibliography{IEEEabrv,bibliography}
%

\end{document}